\documentclass[times,twocolumn,final,authoryear]{article}

\usepackage{framed,multirow}

\usepackage{amssymb}
\usepackage{latexsym}

\usepackage{url}

\begin{document}


\title{On the Intrinsic Limits to Representationally-Adaptive Machine-Learning}

\author{David Windridge \\
{\small Department of Computer Science }\\
{\small School of Science and Technology,  Middlesex University} \\
{\small  The Burroughs, London, NW4 4B, UK}\\
{\small Email: d.windridge@mdx.ac.uk}}


\date{}

\maketitle

\begin{abstract}
Online learning is a familiar problem setting within Machine-Learning in which data is presented serially in time to a learning agent, requiring it to progressively adapt within the constraints of the learning algorithm. More sophisticated variants may involve concepts such as transfer-learning which increase this adaptive capability,  enhancing the learner's cognitive capacities in a manner that can begin to imitate the open-ended learning capabilities of human beings. 

We shall argue in this paper, however, that a full realization of this notion  requires that, in addition to the capacity to adapt to novel data,   autonomous  online learning must ultimately incorporate the capacity to update its  own {\em representational} capabilities  in relation to the data. We therefore enquire about the philosophical  limits of this process, and argue that only fully embodied learners exhibiting an {\em a priori } perception-action link in order to ground representational adaptations are capable of exhibiting the full range of human cognitive capability.

{\bf keywords: philosophy of  machine learning,  perception-action learning,  online learning }

\end{abstract}




\section{Introduction}

In the following, we aim to circumscribe the inherent conceptual limits implicit in the notion of open-ended machine learning - a key criteria for human-like cognition and intelligence - and to propose a strategy for building autonomous agents capable of operating at this extremity of capability. We first commence with a brief statement of the problem. 

\subsection{Conceptual Limits to Open-Ended Learning}

In Putnams's classical `brain in a vat' philosophical thought experiment,  a brain is attached via wires to a supercomputer that simulates all aspects of the real world, mediating this in terms of electrical signals sent down the wires in response to input signals from the brain in the form of nerve impulses. The thought experiment is thus intended to address notions of radical scepticism;  could  such a brain be justified in having true beliefs? (and would these be beliefs about   objects existing within the simulated world or about the input/output characteristics of the electrical signals).

A variant of this thought experiment (in fact a subset of this  thought experiment) might be the notion of a brain in a vat attached, from birth, {\em only} via its optic nerves to a video camera, in front of which pass, in a temporal sequence,  all of the natural scenes of the world. The refined question is then `would such a  brain {\em represent } the world in the same manner that a typical human would; one that is  free to move and interact with the world'?

 It is implicitly argued within this paper that this is not the case; indeed,  it will be argued that such fundamental human perceptual characteristics  as the delineation of the world into discrete objects (i.e. the delineation of entities that are invariant under translation) would not occur to a non-embodied agent.  Any such  relational, non-essentialist notion of representation (\cite{witt01}) has clear implications  for artificial implementations of cognitive learning, which we shall explore in this paper.

In order to evolve this argument, we first consider the problem of how machine  learning takes place in temporalized environments, and address the inherent limitations on this form of learning.

\section{Limits to Standard Approaches to Adaptive Online Learning }

Online learning (\cite{4053654,1315937}) is a standard form of machine-learning induction in which  data is presented serially in time, and in which learning generally takes place one instance at  a time (it is thus the opposite of offline, or batch, learning).  It is also inherently predictive, predicting the label values of data not yet presented to the system.   Such a system is thus inherently {\em adaptive}; the degree of adaptation to new data will vary from on-line learner to on-line learner (sophisticated variants may incorporate notions such as  transfer learning (\cite{pan2010survey,taylor2009transfer}), anomaly detection (\cite{chandola2009anomaly}), and active learning (\cite{settles2010active,koltchinskii2010rademacher})).

Despite this tendency towards increasing adaptivity, however,  the majority of existing approaches  typically assume an underlying consistency in the {\em representational characteristics} of the data; the data-stream presented to an on-line learner  is generally delineated in terms of a fixed  set of classes, or a fixed set of features (for example, spatial interest points or  texture-descriptors). Techniques exist  that partially address these limitations, such as in  online learners that  incorporate Dirichlet processes to spawn novel states in relation to the  requirements of the data (\cite{hoffman2010online}), which are thus capable of expanding their   representational characteristics to a certain  extent. However,  such a learner would not be  capable of spontaneously carrying out as fundamental a data-driven representational shift as that involved in  the transition from, say,  a low-level feature-based representation of the world (delineated e.g. in terms of colored pixels) to an {\em object-based} representation of the world (delineated in terms of indexed entities with associated positions, orientations etc), unless a prior capacity for object representation had been incorporated into it.

Taking the notion of autonomous adaptation to serial data to its conceptual limit would thus  require that both the {\em representational} capabilities of the learner as well its objective knowledge acquisition capabilities should be included in the autonomous learning process. Consider, for example,  the case of an idealized autonomous online learning robot.  Such an idealized online  learner  would thus be capable of spontaneously {\em reparametrizing } its representation of the world in relation to novel sensor data; i.e it must not just be capable of updating it's model  of the world, $W$, generated in terms of some particular representational framework, $R$, (written $R[ W]$), it must also be able to find an appropriate transformation of its representational framework {\it in order to}  `most effectively represent'  the totality of the temporal data, $W$,  via some appropriate criterion. It must thus perform the {\em double} mapping $R[ W] \to R'[ W'] $ (composed of the individual mappings $R \to R' $ and $W \to W' $ ) such that the data $W$ and $W'$ are guaranteed to both represent the same set of entities, as represented by a `noumenal equivalence' predicate $Equiv(R[ W],   R'[ W'])$ or similar (we shall return to this point later).

 In general, `this most effective' representation criterion   will be efficiency based - i.e. we will seek the  mapping $R[ W] \to R'[ W'] $ that minimizes complexity (via e.g an MDL (\cite{rissanen2010minimum}) or Occam's Razor -like criterion). A motivation for this efficiency of representation  criterion can be found in Biosemantics (\cite{mill87}) - humans have adapted over millions of years for efficiency of their representative capability (in terms of either the overall neuronal budget or the total energy of processing). However, there are other aspects to this natural selection of representative capabilities that must be considered (see section (\ref{bio}))

Certain machine learning paradigms are inherently capable of the reparameterization  $R[ W] \to R'[ W'] $, for example, manifold learning techniques (\cite{zhang2012adaptive}) and non-linear dimensionality reduction techniques (\cite{debruyne2010detecting}). The technique adopted is not significant for our wider discussion; the key point is that following the process we arrive at {\em both} a reparameterization framework $R'$ (such as an orthonormal basis in manifold or sub-manifold coordinates) {\em and} a revised data set description  $W'$ in the representational framework (e.g. following projection into the manifold coordinates). Typically, reparameterization will also involve a reduction in the number of parameters required to represent the data - i.e.  the determination of some data-derived sub-manifold $M_s$  necessarily implicates the existence of a projection operator such that the full range of data in the original domain, $W \subset M$, can be mapped  into $M_s$ - for instance, by collapsing data points along the orthogonal complement, $W'^\perp$ ($M$ is thus the original sensory manifold, and $M_s$ the re-mapped representational framework if equipped with a suitable basis). 

Criteria for applying such a reductive reparameterization are many and varied; we might use a stochastically-motivated $2\sigma$ Eigenvalue cut-off in Principle Component Analysis to eliminate noise, for instance, or (more generically), we might use a model selection criterion such as the Akaike information criterion in order to arrive at a principled way to determine the allocation of manifold parameters in relation to the characterization of out-of-model data  (the latter is related to minimum-description length (MDL) approaches, which in turn may be considered approximations of the  `intrinsic' (incomputable) Kolmogorov Complexity of the observed data set).

Whilst there might thus exist an {\it intrinsic} parameterization of any given  dataset when considered only in terms of the efficiency of representation, the ideal choice of representation will also, of necessity,  depend on the {\em purpose} to which the data set is put.  Thus, there will always be meta-reasons for the favoring of a particular data parameterization, a particular representational framework. In fact, efficiency of representation is just such an extrinsic meta-reason; Kolmogorov Complexity is thus not in this sense an intrinsic measure of  the data, but rather a measure driven  by just one among several (potentially infinite) competing requirements for data representation, of which efficiency is only one.

\section{Identity Retention in Online Learning}

However, the above relates to {\it batch} processing of the data, and therefore makes the implicit assumption that all data points are derived from the same source, with perhaps only an instrumentally-irrelevant temporal delay between the collection of data points (that are otherwise independently and identically distributed i.e. they are i.i.d). There is hence a strong assumption of `noumenal continuity' implicit in non-online forms of learning.

This assumption however, becomes complicated when considering an {\em adaptive} online learning, in which both the  data {\em and} the  data representation both have a temporal component. For instance, to give perhaps the simplest instance of this problem, in Simultaneous Location and Mapping (SLAM) robotics (\cite{engelhard2011real,strasdat2010scale}), the robotic agent's model of the world necessarily depends upon its calculation of its own position and orientation in the world (i.e. it must factor its own perspectival world-view into the world model). However, this positional calculation is {\em itself} dependant on (is relative to) the agent's model of the world (i.e. the agent describes its own position and orientation {\em in relation to} the world model). A SLAM agent will therefore position itself in the world (perhaps using active learning  (\cite{fairfield2010active}) in order to minimize model ambiguity) by leveraging its own, uncertain model of the world. Interconnected ambiguities are thus always present in both the agent's self-model (of its location/orientation) and it's model of the world; the hope of SLAM robotics is that, following full exploration of the environment, these ambiguities converge to within some manageable threshold.

In general, the SLAM problem is not soluble unless certain {\em a priori} assumptions are made. A key such assumption is that the environment remains reasonably consistent over time. If an environment were to undergo some arbitrary spatial transformation at each iteration of the SLAM algorithm, then no convergence would be possible (and in fact there would be no meaning to the concept of world model). However, much milder  perturbations of the spatial domain would be sufficient to ensure non-convergence of the algorithm.

A further key {\em a priori} assumption, one that shall be particularly important in the following, but which is often overlooked, relates to the robotic agent's {\em motor capabilities}. The robotic agent's motor capability may, in this case, be considered as {\em that which initiates the change of perspective/change of representation}. However, as such, it cannot in itself be doubted (unlike the world model), and must thus be assumed {\em a priori}. Colloquially, the agent might thus doubt it's location, or its world model, but it cannot, if it is to work at all, doubt the fact that a specific motor impulse has taken place (for instance, a `move forward' or `turn left' command). The agent cannot converge on a world model if, for instance, motor impulses to the actuators underwent arbitrary permutation. Even non-arbitrary permutation would not be distinguishable, even in principle, from a corresponding non-arbitrary permutation of the world space.  (This  non-distinguishability of perceptual manipulations from motor manipulations is absolutely fundamental, and  has important consequences in our later argument).

  Thus,  both the world-model and the agent's (orientation/position-based) self-model are inherently posited {\em relative} its  motor impulses, which can be considered to represent the agent's {\em intentions} in the sense that  the existence of a specific intention  is necessarily not itself open to doubt to the agent, however uncertain its perceptual outcome. Model convergence on a complete world model occurs when the outcome of all actions leads to predictable perceptual consequences (to within some given threshold).  The agent has thus obtained a complete odometery of the environment (in human terms, we have `paced-out' the domain). We can thus consider the world model as being mapped on to a grid of motor impulses such that, in a sense, the agent's active capabilities provide the {\em metric} for its perceptual data (cf also  (\cite{dewe96,glen97,lako99})). 

 In short, where there exists the capacity for updating the representational capacity of an agent in relation to perceptual data that it has sought {\em on the basis of its original representation}, then we need some mechanism for guaranteeing that there is either  sufficient {\em a priori} noumenal knowledge of the external  world, or else sufficient {\em a priori} assumptions made regarding the process (e.g. movement) that initiates new data acquisition,  in order for the representation-updating procedure to converge. Although this is problematic in SLAM, the problem is  much more acute in fully open-ended learning scenarios where whole new {\em categories} of perception can be generated.

Of course, in dealing with  {\em a priori} requirements for perception in an empirical setting, the relevant philosopher is Kant - we now look more closely at this issue in Kantian terms.

\subsection{The Kantian Perspective on Cognitive Agency}

We are essentially, in the above,  asking the question of how, in an adaptive online learning context, is it ever possible for us to empirically validate a proposed change to our representative capability (how is it, in a Popperian sense, possible to {\em falsify} a proposed representational update).  Falsification of a {\em world model} is, by comparison,  straightforward in a standard autonomous robotic system, in that a world model typically constitutes a set of proposed haptic {\em affordances} (\cite{gobs79,mcgr00})  gathered at-a-distance  by a  vision system. Thus, the visual model typically denotes a set of object hypotheses that may be verified via haptic contact  (\cite{saun04,schl03}). 

Haptic contact is thus typically considered to be prior to vision, or at least {\em a priori} less prone to ambiguity than vision. This is also experienced to an extent in  human terms; we tend to consider something that we can touch, but not see (for example, a `force field') as intrinsically having substance, whereas something that we can see  occupying volumetric space but which we cannot verify by touch as being intrinsically  illusory (a holographic image, for instance).

However, in a hypothetical automaton where there exists complete representational fluidity, such that a completely novel sensorium could be developed (for instance by combining sonar data with visual data in some hybrid world description), then we cannot {\em a priori} favor one group of senses/sensors over another in order to delineate  hypotheses about the world.  Moreover, there is no immediately obvious way to form hypotheses about the most appropriate representational framework to adopt. 

In order to address this, we borrow a key insight from Kant; namely that object concepts constitute {\em orderings} of sensory intuitions (\cite{guye99}). Objects, as we understand them do not thus constitute singular percepts, but rather {\em synthetic unities} built upon an {\em a priori} linkage that must be assumed between sensory intuitions and the external noumenal world (these {\em a priori} links cannot be in doubt since the are a {\em condition} of empirical validation for synthetic unities). Implicit in this is the notion that {\em actions} can be deployed to test the validity of these  synthetic unities (which being synthetic rather than analytic are only contingently true, and therefore falsifiable through experience).  Actions are thus causally initiated by the agent and serve to bring aspects of the synthetic unities to attention (within the {\em a priori}  strata of space and time) in a way that renders them falsifiable.

For Kant, assuming that spatiality and temporal causality are  {\em a priori}, means that they are assumed by the agent {\em in order to have falsifiable perceptions} at all; in principle, other ordering approaches to sensory data  may be possible. However, it would be impossible for the agent to retain the continuity and falsifiability of object representation across such a fundamental transition of representation (it would also be impossible for a self-conscious agent  to retain its identity -- or `synthetic unity of apperception'-- across such a fundamental representational chasm). This is the problem of `noumenal continuity' that we identified earlier; how can an agent that undergoes a change of representation framework at time $t_0$  ever be sure that the objects delineated at  $t_0-1$ were the {\em same}  objects as those delineated at  $t_0+1$ (indeed, would the number of objects even be preserved? A cognitive agent might, for instance, hypothesize a perceptual change in which the independent perceptual axes of color-awareness and shape-awareness were combined in a single-dimension unity, such that only one color was allowed per shape, with the corresponding inability to discriminate all of the objects previously discriminated).  An online learner would therefore appear to be severely limited in the extent to which it could utilize data {\em across} representational changes; in short the agent would no longer be a strictly online learner, but  rather a {\em serial batch-learner}. 

However, there is one way in which novel representational changes  can be made while retaining an agent's ability to falsify both these as well as any object hypotheses (synthetic unities) formed {\em in terms of} these  representational changes and, moreover, do so while retaining online continuity of object identity (when extended in perception-action terms -see below). This is when representational changes are built {\em hierarchically}. 

By way of example, consider how, as humans  we typically represent our environment when driving a vehicle. At one level,  we internally represent the immediate   environment in metric-related terms (i.e. we are concerned with our proximity to other road users, to the curb and so on) (\cite{windridge2013characterizing}). At a higher level, however, we are concerned primarily with  {\em navigation}-related entities (i.e how individual roads are {\em connected}). That the latter constitutes a higher hierarchical level, both mathematically and experientially, is guaranteed by the fact  that the topological representation {\em subsumes}, or supervenes upon, the metric representation; i.e. the metric-level provides additional `fine-grained' information to the road topology: the metric representation can be reduced to the topological representation, but not vice versa. In robotics, when goals and  sub-goals are explicitly delineated at each level, this  is known as a {\em subsumption hierarchy} (\cite{broo91}).

In a fully adaptive online learner, it is thus possible to provide a grounded approach to representational induction by adopting a correspondingly  hierarchical approach. Thus, on the assumption of the existence of an {\em a priori} means of validating low-level hypotheses (for example via haptic contact), it is possible to construct falsifiable higher-level representational hypothesis provided that these subsume the latter. Thus, for instance, an embodied autonomous robotic agent might, following active experimentation, spontaneously conceive  a high-level concept of affordance, or  schema (\cite{hintzman1986schema}), such as that of {\em container}. Clearly, in this case, the notion {\em container}  subsumes the concept of  {\em haptic contact}.


Continuity of noumenal identity is thus guaranteed by the lowest level of the hierarchy, with the higher hierarchical levels constituting progressive abstractions and enrichments of the lower level representations. An embodied autonomous robotic age might therefore initially represent the world  in terms of (hypothetical)  volume elements such as voxels or 3d meshes (the  {\em a priori} bootstrap representation), but, following extensive experimentation, might then go on to generate an enriched representation of its world  at a higher level in which containers/and non-containers are  delineated. (Note that the original representation of the world in volumetric terms is thus still present). 

Falsifiability of the representational {\em concept} `container' is thus guaranteed, just as it is possible to guarantee the falsifiability of the   hypothesis   of the existence of any {\em  specific} container, by exploiting the fact that these hypotheses are grounded throughout the hierarchy. Thus, in the former case, the  hypothesis  of the existence of a specific container is rendered falsifiable by haptic contact (and its higher level corollaries); i.e. the agent can test whether the proposed container-entity is, in fact, capable of containing another object.

On the other hand, the  high-level representational \underline{\em concept} `container' is rendered falsifiable by  the fact that it is  conceived along with a corresponding high-level action e.g. `placing an object into a container' which necessarily  subsumes lower-level concepts such as `haptic contact' etc. Thus, the representational  concept is rendered falsifiable on the basis of its  {\em utility} and  {\em compressibility}. 

To see how this works, suppose that an autonomous agent, on discovering by chance exploratory activity (e.g. motor babbling (\cite{modayil2007autonomous})), or via activity driven by lower-level action imperatives, that the previously defined concept `object' yields an exception that allows for objects to be placed co-extantly in the same location as another object (the original  concept `object' assumes that objects placed on top of each other do not then  co-exist in the same location). Further suppose that,  on the basis of this exception, the object class is refined by the agent to accommodate the higher level notion of `container' (i.e. so that the concept of `object' subsumes the concept of `container'). This then constitutes a  {\em  representational hypothesis}, which can be applied to the world (e.g. by training a classifier to distinguish container-objects from non-container-objects). 

The falsifiability of this concept then arises from actively addressing the question whether this higher-level perception of the world (as a series of objects in space that are either container-objects or non-container-objects) in fact constitutes a useful description of the world i.e. whether it yields a net compression in the agent's internal representation of its own possible interactions with the world (its affordances). Thus, if there were only a single container in the world, or if it were not possible to train an accurate classifier for containers in general, then it would be unlikely to constitute a useful description of the world; it would likely be more efficient simply to retain the existing concept of object without modification. However, when the world is in fact constituted of objects for which it is an efficient compression of the agent's  action capability to instigate such a modification of the object concept, then it is appropriate for a representationally-autonomous agent to spontaneously form a higher level of its representational hierarchy. (For an example of this approach utilizing first-order logic induction see (\cite{windridge2010perception})).

 Very often compressibility will be predicated on the discovery of  {\em invariances} in the existing perceptual space with respect to randomized exploratory actions. Thus, for example, an  agent might progress from a pixel-based representation of the world to an object-based representation of the world via the discovery that certain patches of pixels {\em retain their (relative) identity} under translation, i.e. such that it becomes far more efficient to represent the world in terms of indexed objects rather than pixel intensities  (though the latter would, of course, still constitute the base of the  representational hierarchy).   This particular representational enhancement can represent an enormous compression (\cite{wolf87}); a pixel-based representation has a parametric magnitude of $P^n$ (with   $P$ and $n$ being the intensity resolution and number of pixels, respectively), while an object-based representation typically has a parametric magnitude of $\sim n^o$, $o << n$, where $o$ is the number of objects.

In positing this hierarchical approach to representational adaptation, we have thus outlined a framework in which complete representational-autonomy for an embodied machine learner becomes feasible, one in which representations are empirically validatable, and in which  the `noumenal continuity' of identified entities can be assumed across representational transformations. 

A key aspect of this falsifiability is the requirement that the spontaneous generation of higher-level perceptions in the agent's representational hierarchy  correlates directly with  higher level actions. We now look more closely at this perception-action connection, and  consider the low-level  {\em a priori} guarantees of representative falsifiability.


\section{Perception-Action Learning}

Perception-Action learning is a novel paradigm in robotics that aims to address significant deficits in traditional approaches to embodied computer vision (\cite{drey72}). In particular, in the conventional approach to autonomous robotics, a computer vision system will typically be employed to build a model of the agent's environment {\em prior} to the act of planning the agent's actions within the domain. Visual data arising from these actions will then typically be used to further constrain the environment model, either actively or passively (in active learning the agent actions are driven by the imperative of reducing ambiguity in the environment model).

However, it is apparent that there exists in this approach, a very wide disparity between the visual parameterization of the agent's domain and its action capabilities within it (\cite{neha02}). For instance, the parametric freedom of a front-mounted camera will typically encompass the full intensity ranges of the Red, Green and Blue channels of each individual pixel of the camera CCD, such the the range of {\em possible} images that might be generated in each time-frame is of an extremely large order  of magnitude (of course, only a minuscule fraction of this representational space is ever likely to be experienced by the agent). On the other hand, the agent's motor capability is likely to be very much more constrained (perhaps consisting of the possible Euler angle settings of the various actuator motors).  This disparity leads directly to the classical  problems of {\em framing}  (\cite{mcca69}) and {\em symbol grounding} (\cite{harn90}) (note that this observation is not limited purely to vision based approaches - alternative modalities such as LIDAR and SONAR would also exhibit the same issues). 

Perception-Action (P-A) learning aims  to overcome these issues by adopting as its motto, `action {\em precedes} perception' (\cite{gran03,felsberg2009exploratory}). By this it is meant that, in a strict sense (to be defined),  actions are conceptually prior to  perceptions; i.e. that perceptual capabilities should  depend on action capabilities and not vice versa.

Thus, a Perception-Action learning agent proceeds by randomly sampling its action space (`motor babbling'). For each motor action that produces a discernible perceptual output in the bootstrap representation space $S$ (consisting of e.g. camera pixels), a percept $p_i \in S$ is greedily allocated. The agent thus progressively arrives at a set of novel percepts that relate directly to the agent's action capabilities in relation to the constraints of the environment (i.e. the environment's {\em affordances}); the agent learns to perceive only that which it can change. More accurately, the agent learns to perceive only that which it {\em hypothesizes} that it can change  - thus, the set of experimental data points $\cup_i p_i \subset S$ can, in theory, be generalized over so as to create a percept-{\em manifold} that can be mapped onto the action space via e.g. the bijective relation $\{actions \} \to \{ percept_{\rm initial}  \} \times \{ percept_{\rm final} \} $ (i.e. such that each hypothesizable action has a unique, discriminable outcome) \cite{windridge2010perception, wepistem,  windridge2012framework}.


When such a perceptual manifold is created (representing a {\em generalization} over the tested space of action possibilities), this then permits an {\em active} sampling of the perceptual domain - the agent can propose actions with perceptual outcomes that have not yet been experienced by the agent,  but which are consistent with its current representational model (again, this guarantees falsifiability of the perceptual model).  It is in this way that  Perception-Action learning constitutes a form of active learning: randomized selection of perceptual goals within the hypothesized  perception-action manifold leads more rapidly to the capture of data that might falsify the hypothesis than would otherwise be the case (i.e. if the agent were performing randomly-selected actions within in the original motor domain).   Thus, while the system is always 'motor babbling' in a manner analogous to the learning process of infant humans, the fact of carrying out this motor babbling in a higher-level P-A manifold means that the learning  system as a whole  more rapidly converges on the correct model of the world.

Of course, this  P-A motor-babbling activity can take place in {\em any} P-A manifold, of whatever level of abstraction; we may thus,  by combining the idea of P-A learning with the notion of hierarchical representation  presented above, conceive of the notion of a {\em hierarchical  Perception-Action learner} (\cite{shevchenko2009linear}),  in which a vertical representation hierarchy is progressively constructed for which randomized exploratory motor activity at the highest level of the corresponding motor hierarchy would rapidly converge on  an ideal representation of the agent's world in terms of its affordance potentialities. Such a system would thus converge upon both a model of the world, and an ideal strategy for  representation of that world in terms of the learning agent's action capabilities within it.

Perceptual goals thus exist at all levels of the hierarchy, and the subsumptive nature of the hierarchy means that goals and sub-goals are scheduled with increasingly specific  content as the high-level abstract goal is progressively grounded through the hierarchy. (Thus, as humans, we may conceive the high-level intention `drive to work', which in order to be enacted, involves the execution of a large range of sub-goals with correspondingly lower-level perceptual goals e.g. `stay in the center of the lane', etc). 

We finally now look at how such a system for representational updating might have spontaneously evolved in humans, and how the wider question of representational fluidity fits into a biological context.

\section{The Biophilosophical Perspective}
\label{bio}

Biosemantics, as a sub-branch of Biophilosophy, was proposed by  Millikan  (\cite{mill87}) as an attempt to subsume certain philosophical questions of representation and perception  within the purview of biology, and in particular,  the contingencies that arise from consistency with respect to natural selection.

We have indicated earlier  that a key notion of Biosemantics lies in motivating  an efficiency of representation  criterion; organisms  are naturally-selected for efficiency of their representative capability in terms of either overall neuronal budget or total energy of processing.

However, a further aspect implicit in Biosemantics is the {\em embodiment} of the agent. Thus, the biological organism's representative capability must, in addition to being maximally or near-maximally efficient,  also {\em be of utility to the organism} in perpetuating it's genetic code (i.e. it must be consistent with Natural Selection) if it is to be consistently propagated. In practical terms, this means that the organism must be able to discriminate those entities (food, predators, mates etc), that are key to its survival and reproduction (\cite{piag70}).  However, the biological agent will also have acquired, by Natural Selection, an {\it active capability} that is likewise evolved to maximize the organism's ability to  propagate its genetic code; i.e. its ability to interact with the environment is adapted to maximize its survival and reproductive capability; a lobster's claws are evolved for opening shells etc. The perceptual and the active capabilities of most organisms have thus evolved in lock-step; the organism perceives only  (since it must maximize efficiency of representation) that which is relevant to its survival and reproduction in addition to {\it that which it is capable of interacting with} so as to maximize its survival and reproductive capability.

However, this describes a biological  entity with a fixed, evolved representational framework (whether in a natural or simulated  environment (\cite{sipp95})). Humans, however, have, to a larger degree than any other animal acquired the capacity to be able to reconfigure their neuronal and perceptual structure {\em in relation to} the environment in ways that go far beyond the immediate biological requirements. Thus, rather than both the  organism's representational framework, $R$, {\em and} the organism's active capability, $A$,   having being adapted to the world, $w$, over time, in humans beings, the representation framework is capable of adapting  directly to the world, $w$.  (Which is not to say that the {\em possibility} of reconfiguration does not serve  our biological ends, simply that any particular reconfiguration occurs in relation to the biologically-experienced facts of the world, and is not itself naturally-selected for optimality with respect to the organism's long-term capacity survival). These perceptual reconfigurations can be very abstract; we can thus, for instance perceive the world  in terms of the interaction between  socio-economic groups if we are an economist, or in aesthetic terms   if we are an artist.  (Note that we are not suggesting that humans are able to update {\em all} of their representative capacity, only some significant fraction of it).

This reconfigurability of human perceptual structure in relation to  the environment makes it critically different from the more usual naturally-selected perceptual  capability found amongst other organisms. There is, in particular, no immediate survival imperative attached to perceptual reconfiguration, other than by proxy (for instance there may be a constraint on the total neuronal/energy budget involved in  the perceptual reconfiguration). However, such `budgetary'  proxies for the requirements  of natural selection   are not, in themselves, sufficient to motivate any particular  reconfiguration of the human agent's perceptual capability - for  this we need  an additional  proxy criterion, one that leads to a `retention of active capability' (if neuronal efficiency {\em alone} were the criterion for perceptual updating, then it would always be optimal to map increasing numbers of the original percepts on to a singular novel percept). 

The two  principle (non-naturally-selected) operative  criteria for perceptual updating in humans are thus:

\vspace{5pt}

\noindent 1. {\em Obtaining a maximally efficient representation of the environment}

\vspace{5pt}

\noindent in combination with:

\vspace{5pt}

\noindent 2.   {\em Ensuring  the discriminability of the active capabilities of the agent, as well as key entities related to survival/reproduction/nutrition}. 

\vspace{5pt}

By the `discriminability of the active capabilities' in the latter constraint, we mean the ability to perceive the outcomes of intended actions undertaken by the agent i.e. an intentional action (or at least one initiated by the goal-setting aspect of the agent's cognition), should be susceptible to the sensory determination of its having taken place as intended. (In straightforward terms we might say that  an `intentional action' is that which has  a specific percept as its success criterion.)



It is thus clear, whether considered from  an {\em a priori} Kantian, or an {\em a posteriori}  Biophilosophical perspective, that perceptions and actions must retain a fundamental link in any   representationally-adaptive online learning system capable of  emulating human cognitive capabilities.



\section{Conclusion}

We have thus proposed hierarchical perception-action learning as the idealized form of adaptive online-learning, which, by virtue of its embodiment within the environment, is able to empirically validate  both its model of the world {\em and} its  representation of the world.

An important corollary of this approach is that, at no stage, is there any requirement for global hierarchical consistency of representation (thus, as humans, we do not carry around within us a set of exact Cartesian coordinate locations of  the key elements of our native town; rather what we retain is a series of motor imperatives to be triggered in relation to key percepts: e.g. `turn left at the town-hall'). In a sense, in a  Perception-Action learning agent,  ``the environment has become it {\em own} representation'', (\cite{newe76}), which naturally represents a very significant compression of the information that an agent needs to retain.

This relates to the issue of symbol grounding, a seminal problem in the conceptual underpinning of the classical approach to machine learning (\cite{harn90}). The problem arises when one attempts to relate an abstract symbol manipulation system (it was a common historical assumption that computational reasoning would center on first-order logic deduction) with the stochastic, shifting reality of sensor data.   In hierarchical P-A  learning the problem is eliminated by virtue of the fact that representations are {\em abstracted from the bottom-up} (\cite{marr82,gard94,moda04b,gran03}). They are thus always intrinsically grounded (indeed this grounding is the main guarantor of their falsifiability).

We finally  note that motor-babbling at the top of the representation hierarchy would necessarily involve the spontaneous scheduling of perceptual goals and sub-goals at the lower level of the hierarchy in a way that would (as the hierarchy becomes deeper) necessarily look increasingly 'intentional' (a phenomenon that is  readily apparent in  the development of motor movement of human infants). 

Hierarchical P-A learning  would therefore seem  the natural direction of progress in embodied adaptive online learning. The question then arises of how this would apply if the embodiment  that guarantees the falsifiability of representational updating is in  a domain that is not directly physical   e.g. when the adaptive online learner is, for example, a web-crawling robot (indeed, what does `embodiment' mean in this context?).  Could such an agent spontaneously adapt to perceive high level concepts in, for example, html data while retaining the integrity of its underlying `motor space'? 

The answer, in this case, hinges on the fact that the agent's actions are the  {\em searching} and {\em indexing} actions undertaken by the robot; it is embodied in so far as it has a {\em location} with respect to these  action capabilities. At the lowest ({\em a priori}) level there is thus the basic ability to move between web-pages; this capability cannot, under any circumstances be altered by the agent. It is, however, quite free to spontaneously form higher level search and index capabilities {\em built upon these}, for example by meta-indexing documents in terms of discovered higher-level subject-matters. The agent is thus capable of complete flexibility of hierarchical representation with respect to the falsifiability constraints that we have outlined, and is thus a fully-constituted hierarchical P-A learner.

The proposed framework is thus one of very general  applicability, and one which, we believe has the potential to address the  fundamental conceptual deficits in standard notions of adaptive online learning that we have outlined.





\bibliographystyle{IEEEtran}

\bibliography{cogbootbibiog3}

\end{document}